\documentclass[sigconf]{acmart}

\AtBeginDocument{%
  \providecommand\BibTeX{{%
    \normalfont B\kern-0.5em{\scshape i\kern-0.25em b}\kern-0.8em\TeX}}
    }
 
\usepackage{balance}
\usepackage{etoolbox}

\usepackage{diagbox} 
\usepackage{multirow}
\usepackage{nidanfloat}
\usepackage{subcaption}
\usepackage{color, colortbl}
\usepackage{enumerate}   
\definecolor{Gray}{gray}{0.9}
\newcommand{\commentaire}[1]{}

\copyrightyear{2021}
\acmYear{2021}
\setcopyright{acmcopyright}
\acmConference[CIKM '21] {Proceedings of the 30th ACM International Conference on Information and Knowledge Management}{November 1--5, 2021}{Virtual Event, Australia.}
\acmBooktitle{Proceedings of the 30th ACM Int'l Conf. on Information and Knowledge Management (CIKM '21), November 1--5, 2021, Virtual Event, Australia}
\acmPrice{15.00}
\acmISBN{978-1-4503-8446-9/21/11}
\acmDOI{10.1145/3459637.3482321}

\newcounter{BalanceAtReference}
\newcounter{ReferenceIndexForBalancing}

\makeatletter

\global\@ACM@balancefalse

\def\@balancelastpageonce{%
  \ifnum\value{ReferenceIndexForBalancing}=\value{BalanceAtReference}
    \newpage
  \else
    \relax
  \fi
  \stepcounter{ReferenceIndexForBalancing}
}
\pretocmd{\bibitem}{\@balancelastpageonce}
  {} 
  {\@latex@error{Patching \bibitem failed}{\@ehd}}

\makeatother

\begin{document}
\fancyhead{}

\title{ASTERYX: A model-Agnostic SaT-basEd appRoach \\ for sYmbolic and score-based eXplanations \\ (Preprint version)}


\author{Ryma Boumazouza}
\email{boumazouza@cril.fr}
\orcid{0000-0002-3940-8578}
\affiliation{%
  \institution{Univ. Artois, CNRS, CRIL}
  \streetaddress{F-62300}
  \city{Lens}
  \country{France}}

\author{Fahima Cheikh-Alili}
\email{cheikh@cril.univ-artois.fr}
\orcid{0000-0002-4543-625X}
\affiliation{%
  \institution{Univ. Artois, CNRS, CRIL}
  \streetaddress{F-62300}
  \city{Lens}
  \country{France}}

\author{Bertrand Mazure}
\email{bertrand.mazure@univ-artois.fr}
\orcid{0000-0002-3508-123X}
\affiliation{%
  \institution{Univ. Artois, CNRS, CRIL}
  \streetaddress{F-62300}
  \city{Lens}
  \country{France}}
  
\author{Karim Tabia}
\email{karim.tabia@univ-artois.fr }
\orcid{0000-0002-8632-3980}
\affiliation{%
  \institution{Univ. Artois, CNRS, CRIL}
  \streetaddress{F-62300}
  \city{Lens}
  \country{France}} 

\renewcommand{\shortauthors}{R. Boumazouza et al.}

\begin{abstract}
The ever increasing complexity of machine learning techniques used more and more in practice, gives rise to the need to explain  the outcomes of these models, often used as black-boxes. Explainable AI approaches are either numerical feature-based aiming to quantify the contribution of each feature in a prediction or symbolic providing certain forms of symbolic explanations such as \textit{counterfactuals}.
This paper proposes a generic agnostic approach named ASTERYX allowing to generate both symbolic explanations and score-based ones.
 Our approach is declarative and it is based on the encoding of the model to be explained in an equivalent symbolic representation. This latter serves to generate in particular two types of symbolic explanations which are \textit{sufficient reasons} and \textit{counterfactuals}. 
We then associate scores reflecting the relevance of the explanations and the features w.r.t to some properties. 
 Our experimental results show the feasibility of the proposed approach and its effectiveness in providing symbolic and score-based explanations. 
\end{abstract}

\ccsdesc[500]{Computing methodologies~Artificial intelligence}
\ccsdesc[300]{Computing methodologies~Machine learning}
\keywords{XAI;  Symbolic explanations;  Score-based explanation; Model-Agnostic; Satisfiability testing}

\maketitle

\section{Introduction}
In the last decades, the growth of data and widespread usage of Machine Learning (ML) in multiple sensitive fields (e.g. healthcare, criminal justice) and industries emphasized the need for explainability methods. 
 These latter can be grouped into pre-model (ante-hoc), in-model, and post-model (post-hoc). We mainly focus on  post-hoc methods where we distinguish two types of explanations: (1) symbolic explanations (e.g. \cite{shih2018symbolic},\cite{ignatiev2019relating}) that are based on symbolic representations used for explanation, verification and diagnosis purposes (\cite{reiter1987theory},\cite{rymon1994se},\cite{ignatiev2019relating}), and (2) numerical feature-based methods that provide insights into how much each feature contributed to an  outcome (e.g. SHAP\cite{lundberg2017unified}, LIME\cite{ribeiro2016should}). Intuitively, these two categories of approaches try to answer two different types of questions: Symbolic explanations tell why a model predicted a given label for an instance (eg. sufficient reasons) or what would have to be modified in an input instance to have a different outcome (counterfactuals). Numerical approaches, on the other hand, attempt to answer the question to what extent does a feature influence the prediction.

The existing symbolic explainability methods are model-specific (can only be applied to specific models for which they are intended) 
and cannot be applied agnostically to any model, which is their main limitation.  In the other hand,  feature-based methods such as Local Interpretable Model-Agnostic Explanations (LIME)\cite{ribeiro2016should} and  SHapley  Additive  exPlanations (SHAP)\cite{lundberg2017unified} provide the features' importance values for a particular prediction. These values provide an overall information on the contribution of features individually but do not really allow answering certain questions such as: "{\it{}What are the feature values which are sufficient in order to trigger the prediction whatever the values of the other variables?} " or "{\it{}Which values are sufficient to change in the instance $x$ to have a different prediction?}". This type of questions is fundamental for the understanding, and, above all, for
the explanations to be usable. For example, if a user's  application is refused, the user will naturally ask the question: "{\it{}What must be changed in my application to be accepted?} ". We cannot answer this question in a straightforward manner with the features-based explanations. 
 Thus, the major objective of our contribution is to provide  both symbolic explanations and score-based ones for a better understanding and usability of explanations. 
 It is declarative and does not require the implementation of specific algorithms since its based on well-known Boolean satisfiability concepts, allowing to exploit the strengths of modern SAT solvers.
We model our explanation enumeration problem and use modern SAT technologies to enumerate the explanations.  
The approach provides two complementary types of symbolic explanations for the prediction of a data instance $x$: {\bf \textit{Sufficient Reasons}} ({\bf \textit{$SR_x$}} for short) and {\bf \textit{Counterfactuals}} ({\bf \textit{$CF_x$}} for short). In addition, it provides score-based explanations allowing to assess the influence of each feature on the outcome. 
\noindent The main contributions of our paper are :
\begin{enumerate}
    \item A {\bf declarative} and {\bf model-agnostic} approach allowing to provide {\bf \textit{$SR_x$}} and {\bf \textit{$CF_x$}} explanations based on SAT technologies ; 
    \item A set of fine-grained properties allowing to analyze and select explanations and a set of scores allowing to assess the relevance of explanations and features w.r.t the suggested properties ; 
    \item An experimental evaluation providing an evidence of the feasibility and efficiency of the proposed approach ;
\end{enumerate}

\section{Preliminaries and notations}
Let us first formally recall some definitions used in the remainder of this paper. For the sake of simplicity, the presentation is limited to binary classifiers with binary features. 
We explain negative predictions where the outcome is $0$ within the paper but the approach applies similarly\footnote{will be discussed in the "Concluding remarks and discussions" Section.} to explain positive predictions.

\begin{definition}{\textbf{(Binary classifier)}} A Binary classifier is defined by two sets of binary variables: A feature space $X$= \{$X_1$,..,$X_n$\} where $|X|$=$n$, and a binary class variable denoted $Y$.
\end{definition}
A decision function describes the classifier's behavior independently from the way it is implemented. We define it as a function $f:X \rightarrow Y$ mapping each instantiation  $x$ of $X$ to $y$=$f(x)$. A data instance $x$ is the feature vector associated with an instance of interest whose prediction from the ML model is to be explained. We use interchangeably in this paper $f$ to refer to the classifier and its decision function.  
\begin{definition}{\textbf{(SAT : The Boolean Satisfiability problem)}}
Usually called SAT, the Boolean satisfiability problem is the decision problem, which, given a propositional logic formula, determines whether there is an assignment of propositional variables that makes the formula true.
\end{definition}
The logic formulae are built from propositional variables and Boolean connectors "AND" ($\wedge$), "OR" ($\vee$), "NOT" ($\neg$). A formula is satisfiable if there is an assignment of all variables that makes it true. It is said inconsistent or unsatisfiable otherwise.
{For example, the formula {$(x_1\wedge x_2)\vee\neg x_1$} where $x_1$ and $x_2$ are Boolean variables, is satisfiable since if {$x_1$} takes the value false, the formula evaluates to true.} A complete assignment of variables making a formula true is called a model while a complete assignment making it false is called a counter-model. 
 \begin{definition}{\textbf{(CNF (Clausal Normal Form))}} A CNF 
is a set of clauses seen as a conjunction. A clause is a formula composed of a disjunction of literals. A literal is either a Boolean variable $p$ or its negation $\neg p$. A quantifier-free formula is built from atomic formulae using conjunction $\wedge$, disjunction $\vee$, and negation $\neg$. An interpretation $\mu$ assigns values from \{0, 1\} to every Boolean variable. Let $\Sigma$ be a CNF formula, $\mu$ satisfies  $\Sigma$ iff $\mu$ satisfies all clauses of $\Sigma$.
 \end{definition}
  Over the last decade, many achievements have been made to modern SAT solvers\footnote{A SAT solver is a program for deciding the satisfiability of Boolean formulae encoded in conjunctive normal form.} that can handle now problems with several million clauses and variables, allowing them to be efficiently used in many applications. 
 Note that we rely on SAT-solving to explain a black-box model where we encode the problems of generating our symbolic explanations as two common problems related to satisfiability testing which are enumerating {\it minimal reasons} why a formula is inconsistent and {\it minimal changes to a formula} to restore the consistency. Indeed, in the case of an unsatisfiable CNF, we can analyze the inconsistency by enumerating sets of clauses causing the inconsistency (called  Minimal Unsatisfiable Subsets and noted MUS for short), and other sets of clauses allowing to restore its consistency (called Minimal Correction Subsets, MCS for short). The enumeration of MUS/MCS are well-known problems dealt with in many areas such as knowledge base reparation. Several approaches and tools have been proposed in the SAT community  for their generation (e.g. \cite{liffiton2008algorithms,gregoire2007boosting}).

\section{ASTERYX: A global overview }
\begin{figure*}[t]
\centering
  \includegraphics[width=0.75\linewidth]{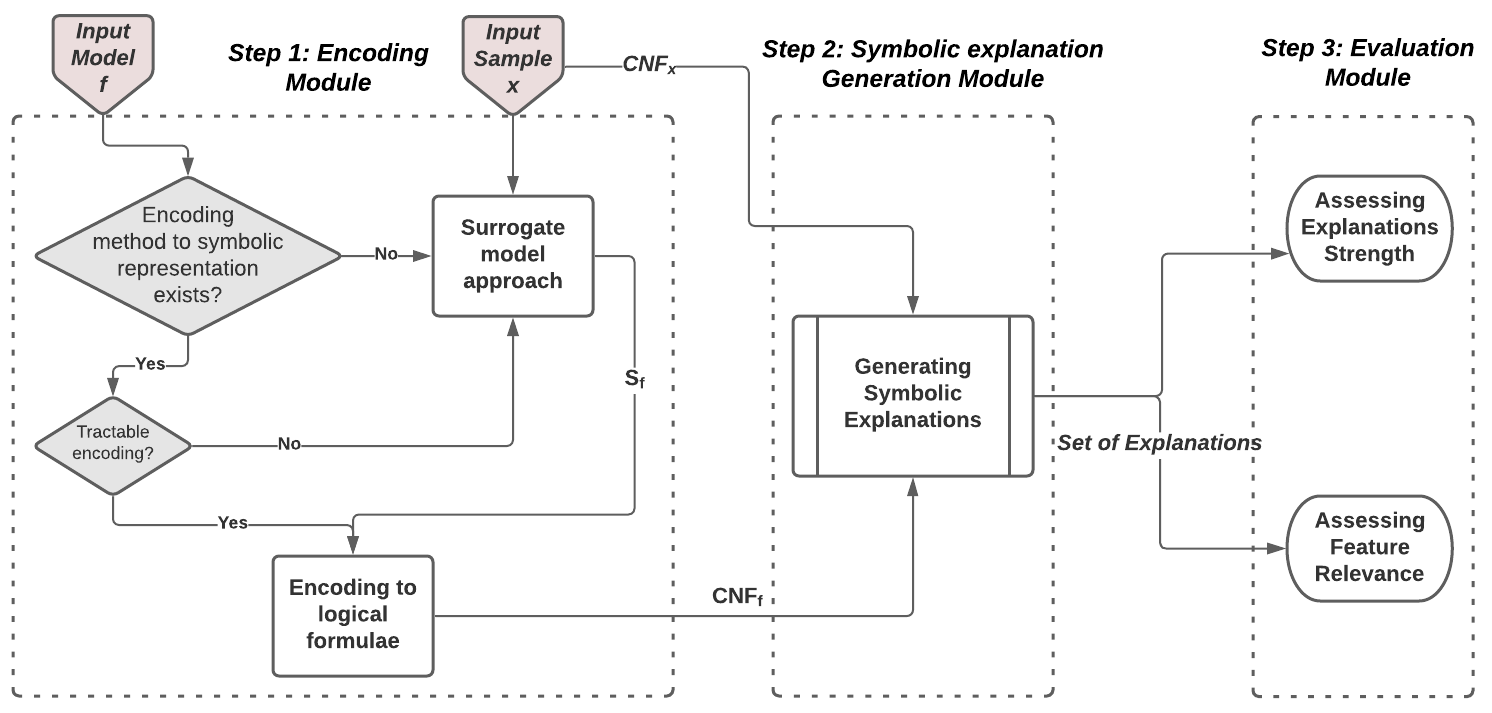} 
\caption{A global overview of the proposed approach}
\label{fig1}
\end{figure*}

Our approach is based on associating a symbolic representation that is (almost) equivalent to the decision function of the model to explain. An overview of our approach is depicted on Figure \ref{fig1}. \\
Given a classifier $f$, our approach proceeds as follows:
\begin{itemize} 
\item {\bf Step 1 (Encoding into CNF the classifier)}: This comes down to associating an {\it equivalent} symbolic representation $\Sigma_f$ to $f$.  $\Sigma_f$ will serve 
to generate symbolic explanations in the next step. The encoding is done either using model encoding algorithms if available and if the encoding is tractable, or using a surrogate approach as described in Section \ref{encodingsec}. 
\item {\bf Step 2 (SAT-based modeling of the explanation enumeration problem)}: Once we have the CNF representation $\Sigma_f$ and the input instance $x$ whose prediction by $f$ is to be explained, we model the explanation generation task as a partial maximum satisfiability problem, also known as Partial Max-SAT \cite{biere2009handbook}. This step, presented in Section \ref{generationsec}, aims to provide two types of symbolic explanations: {\bf \textit{$SR_x$}} and {\bf \textit{$CF_x$}}. They respectively correspond to Minimal Unsatisfiable Subsets (MUS) and Minimal Correction Subsets (MCS) in the SAT terminology. 
\item {\bf Step 3 (Explanation and feature relevance scoring)}: This step aims to assess the relevance of explanations by associating scores evaluating those explanations with regard to a set of properties presented in Section \ref{scoresec}. Moreover, this step allows to assess the relevance of features using scoring functions and to evaluate their individual contributions to the outcome.
\end{itemize} 
The following sections provide insights for each step of our approach.

\section{Encoding the classifier into CNF }\label{encodingsec}  
This corresponds to \textbf{Step 1} in our approach and it aims to encode the input ML model $f$ into CNF in order to use SAT-solving to enumerate our symbolic explanations. 
Two cases are considered: Either an encoding of classifier $f$ into an equivalent symbolic representation exists (non agnostic case), in which case we can use it, or we consider the classifier $f$ as a black-box and we use a surrogate model approach to approximate it in the vicinity of the instance to explain $x$ (agnostic case). 
A direct encoding of the classifier $f$ into CNF is possible for some machine learning models such as Binarized Neural Networks (BNNs) \cite{narodytska2018verifying} and Naive and Latent-Tree Bayesian networks \cite{shih2019compiling}. 
We mainly focus in this paper on the agnostic option used when no direct CNF encoding exists for $f$ or if the encoding is intractable. 

\subsection{Surrogate model encoding into CNF} 
We propose an approach using a surrogate model $f_S$ which is i) as faithful as possible to the initial model $f$ (ensures same predictions) and ii) allows to obtain a tractable CNF encoding. More precisely, we use the surrogate model $f_S$ to approximate the classifier $f$ in the neighborhood of the instance to be explained. Note that one can approximate the classifier $f$ on the whole data set if this latter is available. A machine learning model that can guarantee a good trade-offs between faithfulness and giving a tractable CNF encoding is the one of random forests \cite{ho1995random}. 
As we will see in our experimental study, random forests allow to obtain a good level of faithfulness (in general around 95\%) while giving compact CNF encodings in terms of the number of clauses and variables. Given a data instance $x$ whose prediction by the original model $f$ is to be explained and a data set, we construct the neighborhood of $x$,  noted $V(x,r)$, by sampling data instances within a radius $r$ of $x$. In case the data set is not available, we can draw new perturbed samples around $x$. 
Once the vicinity of $x$ sampled, we train a random forest on the data set composed of ($x_i$, $f(x_i)$) for $i$=$1..p$ where $x_i$ is a sampled data instance, $p$ is the number of sampled instances. Each $x_i$ is labeled with the prediction $f(x_i)$ since the aim is to ensure that the surrogate model $f_S$ is locally (in $x$'s neighborhood) faithful to $f$. 
 \begin{example}\label{exp1}
      As a running example to illustrate the different steps, we trained a Neural Network model $f$ on the United Stated Congressional Voting Records Data Set\footnote{Available at \url{https://archive.ics.uci.edu/ml/datasets/congressional+voting+records}.}. In this example, the label {\it Republican} corresponds to a positive prediction, noted $1$ while the label {\it Democrat} corresponds to a  negative prediction, noted $0$. The trained Neural Network model $f$  achieves 95.74\% accuracy. An input $x$ consists of the following features :\\
      
    \resizebox{0.5\textwidth}{!}{
      {\small 
      \begin{tabular}{c|l} 
      $X_1$& handicapped-infants \\
      $X_2$ & water-project-cost-sharing \\
      $X_3$ & adoption-of-the-budget-resolution \\
      $X_4$ & physician-fee-freeze \\
      $X_5$ & el-salvador-aid \\
      $X_6$ & religious-groups-in-schools \\
      $X_7$ & anti-satellite-test-ban\\
      $X_8$ & aid-to-nicaraguan-contras\\
      \end{tabular} \hspace{.2cm}
       \begin{tabular}{c|l} 
        $X_9$ &mx-missile\\
      $X_{10}$ & immigration \\
      $X_{11}$ &synfuels-corporation-cutback\\
      $X_{12}$ & education-spending\\
      $X_{13}$ & superfund-right-to-sue\\
      $X_{14}$ & crime\\
      $X_{15}$ & duty-free-exports\\
      $X_{16}$ & export-administration-act-south-africa \\
      \end{tabular}}
      
      } \ \\

  Assume an input instance $x$=$(1$,$1$,$1$,$0$,$0$,$0$,$1$,$1$,$1$,$0$,$0$,$0$,$0$,$1$,$0$,$1)$ whose prediction is to be explained. 
  As a surrogate model, we trained a random forest classifier RF$_f$ composed of 3 decision trees (decision tree 1 to 3 from left to right in Fig. \ref{fig:rf}) on the vicinity of the input sample $x$. In this example, RF$_f$ achieved an accuracy of 91.66\% (RF$_f$ is said locally faithful to $f$ as it has a high accuracy in the vicinity of the instance  $x$ to explain). 

\begin{figure}[htp]
\centering
 \resizebox{0.5\textwidth}{!}{
\includegraphics[width=.5\textwidth]{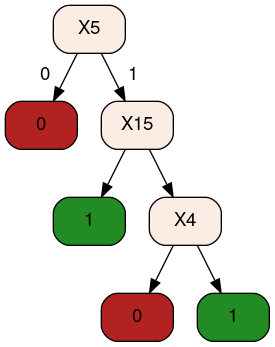}\hfill
\includegraphics[width=.5\textwidth]{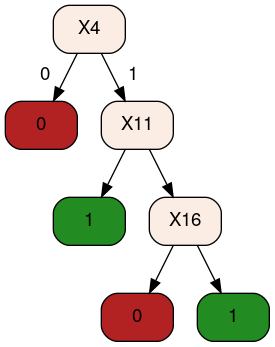}\hfill
\includegraphics[width=.8\textwidth]{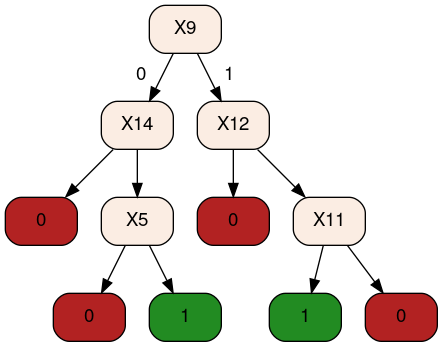}
}
\caption{A random forest trained on the neighborhood of $x$}
\label{fig:rf}

\end{figure}
 \end{example}\vspace{-0.3cm}

The CNF encoding of a classifier $f$ (or its surrogate $f_S$) should guarantee the equivalence of the two representations stated as follows :
\begin{definition}{\textbf{(Equivalence of a classifier and its CNF encoding)}}
A binary classifier $f$ (resp. $f_S$) is said to be equivalently encoded as a CNF $\Sigma_f$ (resp. $\Sigma_{f_S}$) if the following condition is fulfilled: 
$f(x)$=$1$ (resp. $f_S(x)$=$1$) iff $x$ is a model of $\Sigma_f$ (resp. $\Sigma_{f_S}$).
\end{definition}
Namely, data instances $x$  predicted positively ($f(x)$=$1$) by the classifier are models of the CNF encoding the classifier. Similarly, data instances $x$ predicted negatively ($f(x)$=$0$) are counter-models of the CNF encoding the classifier.  

\subsection{CNF encoding of random forests}
When we adopt the surrogate approach and use a random forest $f_S$ to agnostically approximate a classifier $f$, encoding the random forest in CNF amounts to encoding the decision trees individually and then encoding the combination rule (majority voting rule). \\
\noindent {\bf - Encode in CNF every decision tree:} The internal nodes of a decision tree $ DT_i $ represent a binary test on one of the features\footnote{Remember that all the features in our case are binary.}.  
The leaves of a decision tree, each is annotated with the predicted class (namely, $0$ or $1$). A decision tree in our case represents a Boolean function. 

As shown on Example \ref{exp2}, 
the Boolean function encoded by a decision tree can be captured in CNF as the conjunction of the negation of paths leading from the root node to leaves labelled $0$.\\ 
\noindent{\bf  - Encode in CNF the combination rule:} Let $y_i$ be a Boolean variable capturing the truth value of the CNF associated to a decision tree $DT_i$. Hence, the majority rule used in random forests to combine the predictions of $m$ decision trees can be seen as a cardinality constraint\footnote{In our case this constraint means that at least $t$ decision trees predicted the label $1$.} \cite{sinz2005towards} that can be stated as follows:
\begin{equation}
y \Leftrightarrow \sum_{i=1..m} y_i \geq t,
\end{equation}
where $t$ is a threshold (usually $t$=$\frac{m}{2}$). Cardinality constraints have many CNF encodings (e.g. \cite{sinz2005towards,bailleux2003efficient,abio2013parametric}). 
{\color{black}To form the CNF corresponding to the entire random forest, it suffices to conjuct the $m$ CNFs associated to the equivalences between $y_i$ and the CNF of the decisions trees, and, the CNF of the combination rule.} 

\begin{example}[Example \ref{exp1} continued]\label{exp2}
 Let us continue with the random forest classifier of Example \ref{exp1}. The following formulae 
 illustrate the encoding steps applied to $RF_f$:\\
 
 {\small
 \begin{tabular}{lrlll}
\rowcolor{Gray} $DT_1$ & $\!\!\!\!\!y_1$ $\Leftrightarrow$ & $\!\!\!\!\!(X_5)$ $\wedge$ \\
\rowcolor{Gray} & & $\!\!\!\!\!(\neg  X_5 \vee \neg X_{15}\vee X_{4})$ \\
  & &\\
\rowcolor{Gray} $DT_2$ & $\!\!\!\!\!y_2$ $\Leftrightarrow$ & $\!\!\!\!\!(X_4)$  $\wedge$ \\
 \rowcolor{Gray} & & $\!\!\!\!\!(\neg X_4 \vee \neg X_{11} \vee X_{16})$\\
   & & \\
\rowcolor{Gray}  $DT_3$ & $\!\!\!\!\!y_3$ $\Leftrightarrow$ &  $\!\!\!\!\!(X_9 \vee X_{14})$ $\wedge$ \\
\rowcolor{Gray}  &&  $\!\!\!\!\!(X_9 \vee \neg X_{14} \vee X_5)$ $\wedge$ \\
 \rowcolor{Gray}  & & $\!\!\!\!\!(\neg X_9 \vee X_{12})$ $\wedge$\\
 \rowcolor{Gray}  && $\!\!\!\!\!(\neg X_9 \vee \neg X_{12} \vee \neg X_{11})$ \\
   & & \\
\rowcolor{Gray} Majority vote & $\!\!\!\!\!y$  $\Leftrightarrow$ & $\!\!\!\!\!(y_1$$\wedge$$y_2)$ $\vee$ $(y_1$$\wedge$$y_3)$ $\vee$ $(y_2$$\wedge$$y_3)$ $\vee$ $(y_1$$\wedge$$y_2$$\wedge$$y_3)$\\ 
 \end{tabular}
 
}\ \\

In this example, each decision tree ($DT_i$, $i$=$1..3$) represents a Boolean function whose truth value is captured by Boolean variable $y_i$. The random forest $RF_f$ Boolean function is captured by the variable $y$.  Note that the encoding of $RF_f$  is provided in this example in propositional logic in order to avoid heavy notations. Direct encoding to CNF could easily be obtained  using for example Tseitin Transformation \cite{tseitin1983complexity}.
\end{example}

\section{Generating Sufficient Reasons and Counterfactual explanations}\label{generationsec}
In this section, we present {\bf \textit{$SR_x$}} and {\bf \textit{$CF_x$}} as well as the SAT-based setting we use to generate such explanations where the input is the CNF encoding of a classifier $\Sigma_f$ and an input data instance $\Sigma_x$ whose prediction is to be explained.

\subsection{STEP 2: A SAT-based setting for the enumeration of symbolic explanations}
Recall that we are interested in two complementary types of symbolic explanations: the {\it sufficient reasons} ({\bf \textit{$SR_x$}}) which lead to a given prediction and the {\it counterfactuals} ({\bf \textit{$CF_x$}}) allowing to know minimal changes to apply on the data instance $x$ to obtain a different outcome. Our approach to enumerate these two types of explanations is based on two very common concepts in SAT which are MUS and MCS that we will define formally in the following. To restrict the explanations only to clauses that concern the input data $x$ and do not include clauses that concern the encoding of the classifier, we use a variant of the SAT problem called Partial-Max SAT \cite{biere2009handbook} which can be efficiently solved by the existing tools implementing the enumeration of MUSes and MCSes such as the tool in \cite{gregoire2018boosting}. 

A Partial Max-SAT problem is composed of two disjoint sets of clauses 
where $\Sigma_H$  denotes the hard clauses (those that could not be relaxed) and $\Sigma_S$ denotes the soft ones (those that could be relaxed). In our modeling, the set of hard clauses corresponds to $\Sigma_f$ and the soft clauses to $\Sigma_x$ representing the CNF encoding of the data instance $x$ whose prediction $f(x)$ is to be explained. Let $\Sigma_x$ be the {\it soft clauses}, defined as follows : 
\begin{itemize}
    \item Each clause $\alpha \in \Sigma_x$ is composed of exactly one literal ($\forall \alpha \in \Sigma_x, |\alpha|= 1$).
    \item Each literal representing a Boolean variable of $\Sigma_x$ corresponds to a Boolean variable $\{ X_i \in X\}$. 
\end{itemize}
Recall that since the classifier $f$ is equivalently encoded to $\Sigma_f$, then a negative prediction $f(x)$=$0$ corresponds to an unsatisfiable CNF $\Sigma_f$$\cup$$\Sigma_x$. Now, given an unsatisfiable CNF $\Sigma_f$$\cup$$\Sigma_x$, it is possible to identify the subsets of $\Sigma_x$ responsible for the unsatisfiability (corresponding to  reasons of the prediction $f(x)$=$0$), or the ones allowing to restore the consistency of $\Sigma_f$$\cup$$\Sigma_x$ (corresponding to \textit{counterfactuals} allowing to flip the prediction and get $f(x)$=$1$). 
\subsection{Sufficient Reason Explanations ($SR_x$)}
We are interested here in identifying minimal reasons why the prediction is $f(x)$=0. This is done by identifying subsets of clauses causing the inconsistency of the CNF $\Sigma_f$$\cup$$\Sigma_x$ (recall that the prediction $f(x)$ is captured by the truth value of $\Sigma_f$$\cup$$\Sigma_x$). Such subsets of clauses encoding the input $x$ are \textit{sufficient reasons} for the prediction being negative. We formally define the {\bf \textit{SR$_x$}} explanations as follow:
\begin{definition}{\textbf{({\bf \textit{SR$_x$}} explanations)}}\label{sr}
Let $x$ be a data instance and $f(x)$=$0$ its prediction by the classifier $f$. A sufficient reason explanation $\textit{\~{x}}$ of $x$ is such that:
\begin{enumerate}[i.]
    \item\label{itemi} $\textit{\~{x}}$ $\subseteq x$ ($\textit{\~{x}}$ is a part of $x$)
    \item\label{itemii} $\forall \acute{x}$, $\textit{\~{x}}$ $\subset \acute{x}:$ $f(\acute{x})$=$f(x)$  ($\textit{\~{x}}$ suffices to trigger the prediction)
    \item There is no partial instance $\hat{{x}} \subset \textit{\~{x}}$ satisfying \ref{itemi} and \ref{itemii} (minimality)
\end{enumerate}
\end{definition}

\noindent Intuitively, a \textit{sufficient reason} $\textit{\~{x}}$ is defined as the part of the data instance $x$ such that  $\textit{\~{x}}$ is minimal and causes the prediction $f(x)$=$0$. 
We now define Minimal Unsatisfiable Subsets :
\begin{definition}{(\textbf{MUS})}
A Minimal Unsatisfiable Subset (MUS) is a minimal subset $\Gamma$ of clauses of a CNF $\Sigma$ such that $\forall$ $\alpha$ $\in$ $\Gamma$, $\Gamma \backslash \{\alpha\}$ is
satisfiable. 
\end{definition}
\noindent Clearly, a MUS for $\Sigma_f$$\cup$$\Sigma_x$ comes down to a subset of soft clauses, namely a part of $x$ that is causing the inconsistency, hence the prediction $f(x)$=0. 

\begin{proposition}\label{srmuc}
Let $f$ be a classifier, let $\Sigma_f$ be its CNF representation. Let also $x$ be a data instance predicted negatively ($f(x)$=$0$) and let $\Sigma_f$$\cup$$\Sigma_x$ be the corresponding Partial Max-SAT encoding.   Let $SR(x,f)$ be the set of \textit{sufficient reasons} of $x$ wrt. $f$.
Let MUS$(\Sigma_{f,x})$ be the set of MUSes of $\Sigma_f$$\cup$$\Sigma_x$.
Then: 
\begin{equation}\label{musisrsiff}
        \forall \text{\~{x}} \subseteq x, \text{\~{x}} \in SR(x,f)   \iff \text{\~{x}} \in MUS(\Sigma_{f,x})
\end{equation}
\end{proposition}
\noindent Proposition \ref{srmuc} states that each MUS of the CNF $\Sigma_f$$\cup$$\Sigma_x$ is a $SR_x$ for the prediction $f(x)$=0 and vice versa. The proof is straightforward. It suffices to remember that the decision function of $f$ is equivalently encoded by $\Sigma_f$ and that the definition of a MUS on $\Sigma_f$$\cup$$\Sigma_x$ corresponds exactly to the definition of an $SR_x$ for $f(x)$.

\begin{example}[Example \ref{exp2} continued]\label{exp3}
     Given the CNF $\Sigma_f$$\cup$$\Sigma_x$ associated to $RF_f$ from Example \ref{exp2} and the input $x$=$(1,1,1,0,0,0,1,1,1,$ $0,0,0,0,1,0,1)$, we enumerate the {\bf \textit{$SR_x$}} for $f(x)$=$0$ ($x$ is predicted as {\it Democrat}). There are three {\bf \textit{$SR_x$}}:
    \begin{itemize}
        \item  $SR_x1$={"$X_4$=$0$ AND $X_5$=$0$"} (meaning that if the features {\it physician-fee-freeze ($X_4$) } and {\it el-salvador-aid ($X_5$)}  are set to 0, then the prediction is $0$) ;
        \item  $SR_x2$={"$X_{12}$=$0$ AND $X_5$=$0$" } ;
        \item  $SR_x3$={"$X_{4}$=$0$ AND $X_{12}$=$0$ AND $X_9$=$1$"} ;
    \end{itemize}
    
It is easy to check for instance that if $X_4$=$0$ and $X_5$=$0$ then $DT_1$ and $DT_2$ of Fig. \ref{fig:rf} predict $0$ leading the random forest to predict $0$. 
\end{example}

\subsection{Counterfactual Explanations ($CF_x$)}
For many applications, knowing the reasons for a prediction is not enough, and one may  need to know what changes in the input need to be made to get an alternative outcome. Let us formally define the concept of counterfactual explanation.
\begin{definition}{({\bf \textit{CF$_x$}} Explanations)}\label{ce}
Let $x$ be a complete data instance and $f(x)$ its prediction by the decision function of $f$. A \textit{counterfactual} explanation $\textit{\~{x}}$ of $x$ is such that:
\begin{enumerate}[i.]
    \item $\textit{\~{x}} \subseteq x$ ($\textit{\~{x}}$ is a part of x)
    \item $f(x[\textit{\~{x}}])$= 1-$f(x)$ (prediction inversion)
    \item There is no $\hat{{x}}$ $\subset$ $\textit{\~{x}}$ such that $f(x[\hat{{x}}])$=$f(x[\textit{\~{x}}])$ (minimality)
\end{enumerate}
\end{definition}
\noindent In definition \ref{ce}, the term $x[\textit{\~{x}}]$ denotes the data instance $x$ where variables included in $\textit{\~{x}}$ are inverted. 
In our approach, {\bf \textit{CF$_x$}} are enumerated thanks to the Minimal Correction Subset enumeration\cite{gregoire2018boosting}. 

\begin{definition}{\textbf{(MSS)}}
A Maximal Satisfiable Subset (MSS) $\Phi$ of a CNF $\Sigma$ is a subset (of clauses) $\Phi$ $\subseteq$ $\Sigma$ that is satisfiable and such that $\forall$ $\alpha$ $\in$ $\Sigma$ $\backslash$ $\Phi$, $\Phi$ $\cup$ \{$\alpha$\} is unsatisfiable.
\end{definition}

\begin{definition}{\textbf{(MCS)}}
A Minimal Correction Subset $\Psi$ of a CNF $\Sigma$ is a set of formulas $\Psi$ $\subseteq$ $\Sigma$ whose complement in $\Sigma$, i.e., $\Sigma$ $\backslash$ $\Psi$, is a maximal satisfiable subset of $\Sigma$.
\end{definition}
\noindent Following our modeling, an MCS for $\Sigma_f$$\cup$$\Sigma_x$ comes down to a subset of soft clauses denoted $\textit{\~{x}}$, namely a part of $x$ that is enough to remove (or reverse) in order to restore the consistency, hence to flip the prediction $f(x)$=$0$ to $f(x[\textit{\~{x}}])$=$1$.
\begin{proposition}\label{cemsc}
Let $f$ be the decision function of the classifier, let $\Sigma_f$ be its CNF representation. Let also $x$ be a data instance predicted negatively ($f(x)=0$) and $\Sigma_f$$\cup\Sigma_x$ the corresponding Partial Max-SAT encoding.
Let $CF(x,f)$  be the set of counterfactuals of $x$ wrt. $f$.
Let MCS$(\Sigma_{f,x})$ the set of MCSs of $\Sigma_f\cup\Sigma_x$. 
Then:  
\begin{equation}\label{mcsiscfiff}
        \forall \textit{\~{x}} \subseteq x, \textit{\~{x}} \in CF(x,f)   \iff \textit{\~{x}} \in MCS(\Sigma_{f,x})
\end{equation}
\end{proposition}

\noindent Proposition \ref{cemsc} states that each MCS of the CNF $\Sigma_f$$\cup$$\Sigma_x$ represents a $CF$ $\textit{\~{x}}$$\subseteq$$x$ for the prediction $f(x)$=0 and vice versa. 

\begin{example}[Example \ref{exp3} cont'd]\label{exp4}
     Given the CNF $\Sigma_f$$\cup$$\Sigma_x$ associated to $RF_f$ from Example \ref{exp2} and the input $x$=$(1,1,1,0,0,0,1,1,1,0,$ $0,0,0,1,0,1)$, we enumerate the counterfactual explanations to identify the minimal changes to alter the vote\textit{Democrat} to \textit{Republican}. 
    There are four {\bf \textit{CF$_x$}}: 
    \begin{itemize}
        \item $CF_x1$={"$X_{4}$=$0$ AND $X_{12}$=$0$"} (meaning that in order to force the prediction to be $1$, it is enough to alter $x$ by setting only the variables {\it physician-fee-freeze ($X_4$) } and {\it education-spending ($X_{12}$)} to 1 while keeping the remaining values unchanged);
        
        \item $CF_x2$={"$X_{5}$=$0$ AND $X_{12}$=$0$"} ;
       
       \item $CF_x3$={"$X_{5}$=$0$ AND $X_9$=$1$"} ;
       
       \item $CF_x4$={"$X_{4}$=$0$ AND $X_{5}$=$0$"} ;
    \end{itemize}
   
It is easy to see that the four {\bf \textit{CF$_x$}} allow to flip the negative prediction associated to $x$. Indeed, in Fig. \ref{fig:rfex}, the pink lines show the branches of the trees that are fixed by the current input instance $x$. Clearly,  according to $CF_x1$="$X_{4}$=$0$ AND $X_{12}$=$0$", if we set $X_{4}$=$1$ and $X_{12}$=$1$ then this will force $DT_2$ and $DT_3$ to predict $1$ making the prediction of the random forest flip to $1$. 
\begin{figure}[htp]
\centering
 \resizebox{0.5\textwidth}{!}{
\includegraphics[width=.4\textwidth]{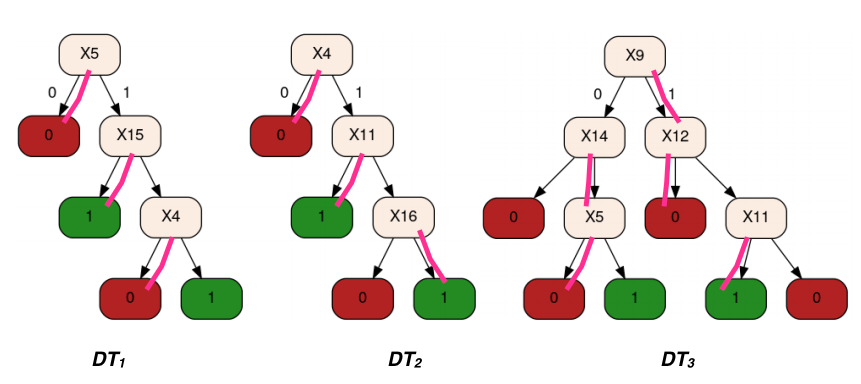}\hfill
}
\caption{The random forest paths set by $x$}
\label{fig:rfex}

\end{figure}
\end{example}

Until now, we presented {\bf Step 1} allowing to encode in CNF a classifier $f$ and {\bf Step 2} allowing to enumerate symbolic explanations that are sufficient reasons and counterfactuals. There remains to numerically assess the relevance of such explanations on the one hand and assess the contribution to the prediction of each feature individually on the other hand. This is the objective of {\bf Step 3} presented in the following section. 

\section{Numerically assessing the relevance of symbolic explanations and features}\label{scoresec}

The number of symbolic explanations from {\bf Step 2} can be large and a question then arises which explanations to choose or which explanations are most relevant?\footnote{An inconsistent Boolean formula can potentially have a large set of explanations (MUSes and MCSes). More precisely, for a knowledge base containing $k$ clauses, the number of MUSes and MCSes can be in the worst case exponential in $k$ \cite{liffiton2008algorithms}.}   
 We try to answer this question by defining some desired properties of an explanation score. \\
 Hence, in order to select the most relevant\footnote{Of course, the relevance depends on the user's interpretation and the context.} explanations and features, we propose to use some natural properties and propose some examples of scoring functions to assign a numerical score to an explanation and to a feature value of the input data.  
 
\subsection{Properties of symbolic explanations and scoring functions}\label{expprop}
 Let us use $E(x,f)$ to denote the set of explanations (either {\bf \textit{SR$_x$}} or {\bf \textit{CF$_x$}}) for an input instance $x$ predicted negatively by the classifier $f$. An explanation is denoted by $e_i$ where $i=1,..,|E(x,f)|$ and $E(x,f)$ is a non empty set. 
The neighborhood of $x$ within the radius $r$ is formally defined as $V(x,r)$: $\{v$$\in$ $X$ $\mid$ diff$(x,v)$ $\leqslant$$r\}$\footnote{diff(x,v) denotes a distance measure that returns the number of different feature values between $x$ and $v$.}. Given an explanation $e_i$, let \textit{size($e_i$)} denote the number of variables composing it, and \textit{Extent($e_i,x,r$)} be the set of data instances defined as : \{$v$$\in$$V(x,r)$ $\mid$ $f(v)$=$f(x)$ and for $e_i$$\in$$E(v,f)$\}. Intuitively, \textit{Extent($e_i,x,r$)} denotes the set of data instances from the  neighborhood of $x$ that are negatively predicted by $f$ and sharing the explanation $e_i$.\\
In the following we propose three natural properties that can be used to capture some aspects of our symbolic explanations :\\
 {\bf - Parsimony ($\mathcal{PAR}$)} : The parsimony is a natural property allowing to select the simplest or shortest explanations (namely, explanations involving less features). Hence, the parsimony score of an explanation $e_i$ should be inversely proportional to it's size. \\
   Formally, given a data instance $x$, its set of explanations $E(x,f)$ : 
       For two explanations $e_1$ and $e_2$ from $E(x,f)$: $\mathcal{PAR}$($e_1$)$>\mathcal{PAR}$($e_2$) iff size($e_1$)$<$size($e_2$) . 
    An example of a scoring function satisfying the parsimony property is :
       \begin{equation}
       S_{\mathcal{PAR}}(e_i) = \frac{1}{size(e_i)}  
   \end{equation}
 
\noindent {\bf - Generality ($\mathcal{GEN}$)} : This property aims to reflect  how much an explanation can be general  to a multitude of data instances, or  in the opposite, reflect  how much an explanation is specific to the instance. Intuitively, the generality of an explanation should be  proportional to the number of data instances it explains. Given a data instance $x$, its set of explanations $E(x,f)$, its neighborhood $V(x,r)$ and two explanations $e_1$ and $e_2$ from $E(x,f)$: 
  $\mathcal{GEN}$($e_1$)$>\mathcal{GEN}$($e_2$)  iff |\textit{Extent($e_1,x,r$)}|$>$|\textit{Extent($e_2,x,r$)}|.   
An example of a scoring function capturing this property is :
 \begin{equation}
         S_{\mathcal{GEN}}(x,r,e_i)= \frac{|Extent(e_i,x,r)|}{|V(x,r)|}
    \end{equation}

Intuitively, this scoring function assesses the proportion of data instances in the neighborhood of the instance $x$ that are negatively predicted and that share the explanation $e_i$. \\

\noindent {\bf - Explanation responsibility ($\mathcal{RESP}$)} : This property allows to answer the question how much an explanation is responsible for the current prediction. Intuitively, if there is a unique explanation, then this latter is fully responsible. Hence, the responsibility of an explanation should be inversely proportional to the number of  explanations in  $E(x,f)$. Given two different data instances $x_1$ and $x_2$ and their explanation sets $E(x_1,f)$ and $E(x_2,f)$ respectively and $e_k \in E(x_1,f)$$\cap$$E(x_2,f)$:  \\
$\mathcal{RESP}$($x_1,e_k$)$<\mathcal{RESP}$($x_2,e_k$)  iff $|E(x_1,f)|$$>$$|E(x_2,f)|$. 
For a given data instance $x$, the responsibility of $e_i$$\in$$E(x,f)$ could be evaluated using the following scoring function : 
\begin{equation}\label{resp}
        S_{\mathcal{RESP}}(x,e_i)= \frac{1}{|E(x,f)|}
   \end{equation}
Note that the scoring function of Eq. \ref{resp} assigns the same score to every explanation in $E(x,f)$. To decide among the explanations in $E(x,f)$, one can calculate a responsibility score for $e_i$ in the neighborhood of $x$. An example of a scoring function capturing this property, would be :
\begin{equation}
        S_{\mathcal{RESP}}(x,r,e_i) = \max\underset{v \in V(x,r) \mid e_i \in E(v,f)} {(S_{\mathcal{RESP}}(v,e_i))}
    \end{equation}

  These properties make it possible to analyze and if necessary select or order the symbolic explanations according to a particular property. Of course, we can define other properties or variants of these properties (e.g. relative parsimony to reflect the parsimony of one explanation compared to the parsimony of the rest of the explanations). The properties can have a particular meaning or a usefulness depending on the applications and users. It would be interesting to study the links and the interdependence between these properties.
 Let us now see properties allowing to assess the relevance of the features reflecting their contribution to the prediction.
\subsection{Properties of features-based explanations and scoring functions}
Let us define \textit{Cover($X_k$,$x$)} as the set of explanations from $E($x$,f)$ where the feature $X_k$ is involved (namely \textit{Cover($X_k$,$x$)}=$\{e_i$$\mid$$X_k$$\in$$e_i$ for $e_i$$\in$$E(x,f)\}$).
We consider the following properties : \\
 {\bf - Feature Involvement ($\mathcal{FI}$)} : This property is intended to reflect the extent of involvement of a feature within the set of explanations. The intuition is that a feature that participates in several explanations of the same instance $x$ should have a higher importance compared to a less involved feature. 
 Given a data instance $x$, its set of explanations $E(x,f)$, and two features $X_1$ and $X_2$:\\
        $\mathcal{FI}$($X_1$,$x$)$>\mathcal{FI}$($X_2$,$x$) iff |\textit{Cover($X_1$,$x$)}|$>$|\textit{Cover($X_2$,$x$)}|.
    An example of a scoring function capturing this property is :
    \begin{equation}
    S_{\mathcal{FI}}(X_k,x)= \frac{|Cover(X_k,x)|}{|E(x,f)|}
    \end{equation}
   
    {\noindent \bf - Feature Generality ($\mathcal{FG}$)} : 
     This property captures at what extent a feature is frequently involved in explaining instances in the vicinity of the sample to explain. 
     Given a sample $x$, its vicinity $V(x,r)$ and the explanation set $E(V(x,r),f)$ defined as  $\underset{v\in V(x,r)}{\bigcup E(v,f)}$, we have: \\ 
    $\mathcal{FG}$($X_1$)$>\mathcal{FG}$($X_2$) iff $\underset{v \in V(x,r)}{|\bigcup Cover(X_1,v)|}$ $>\underset{v \in V(x,r)}{|\bigcup Cover(X_2,v)|}$.  
    An example of a scoring function capturing this property could be :
        \begin{equation}
            S_{\mathcal{FG}}(X_k)=  \frac{\underset{v \in V(x,r)}{|\bigcup Cover(X_k,v)|}}{|E(V(x,r),f)|}
        \end{equation}
    {\noindent \bf - Feature Responsibility ($\mathcal{FR}$)} : This property is intended to reflect the responsibility or contribution of a feature $X_i$ within the set of symbolic explanations of $x$. Intuitively, the responsibility of a feature should be inversely proportional to the size of the explanations where it is involved (the shortest the explanation, the highest the responsibility value of its variables). 
 Given two features  $X_1$, $X_2$ with non empty covers: \\
    $\mathcal{FR}$($X_1$)$>\mathcal{FR}$($X_2$) iff $\underset{e_j \in Cover(X_1,x)}{aggr(size(e_j))}$ $<$ $\underset{e_j \in Cover(X_2,x)}{aggr(size(e_j))}$ where $aggr$ stands for an aggregation function (e.g. $\min$, $\max$, $AVG$, etc.). An example of a scoring function satisfying this property is :
     \begin{equation}
            S_{\mathcal{FR}}(X_k)=\frac{1}{\underset{e_j \in Cover(X_k,x)}{AVG(size(e_j))}}
        \end{equation}

Note that this is a non-exhaustive list of properties that one could be interested in order to select and rank explanations or features according to their contributions. In addition to the different explanation scores presented above, one can aggregate them (e.g., by averaging) to get an overall score depending on the user needs. 
\section{Empirical evaluation}
\subsection{Experimentation set-up}
We evaluated our approach on a widely used standard ML dataset: the MNIST \footnote{\url{http://yann.lecun.com/exdb/mnist/}} handwritten digit database composed of 70,000 images of size 28 × 28 pixels. The images were binarized using a threshold $T=127$. 
In addition, we used three other publicly available datasets (SPECT, MONKS and Breast-cancer). 
We trained \textit{"one-vs-all"} binary neural network (BNN)\footnote{defined as a neural networks with binary weights and activations at run-time} classifiers on the MNIST database to recognize digits (0 to 9) using the pytorch implementation\footnote{available at: https://github.com/itayhubara/BinaryNet.pytorch} of the Binary-Backpropagation algorithm BinaryNets \cite{NIPS2016_d8330f85}. Neural network classifiers were trained on the rest of the datasets. Those classifiers are considered as the input black-box models we are interested in explaining their outcomes.

All experiments have been conducted on Intel Core i7-7700
(3.60GHz ×8) processors with 32Gb memory on Linux. 
\begin{table*}[htb!]
\scriptsize
       \resizebox{\textwidth}{!}{ \begin{tabular}{p{1.75cm}|p{1.cm}|p{1.cm}|p{1.cm}|p{1.cm}|p{1.cm}|p{1.cm}|p{1.cm}|p{1.cm}|p{1.cm}|}
        \hline
          & MNIST\_0&MNIST\_2  & MNIST\_5 & MNIST\_6& MNIST\_8 & SPECT & MONKS&Breast\_cancer \\  
          \hline
         avg acc of RF &{98\%}&93\%&99\%& 96\%& 95\% &99\%&98\%&82\%\\
          \hline
         min size CNF &1744/4944& 1941/5452& 2196/6102& 1978/5534& 1837/5178 &2495/7174&2351/6714&5094/14184 \\
          \hline
         avg size CNF &1979/5540& 2172/6050& 2481/6856 & 2270/6293 &2059/5727 &2758/7921&2883/8146&6069/16907\\
          \hline
          max size CNF &2176/6066& 2429/6760& 2789/7694& 2558/7028& 2330/6408 &3088/8844&3451/9694&7053/19586\\
          \hline
          min enc\_runtime (s)&0.83& 0.88& 0.92& 0.82& 0.74 &1.07&1.66&2.02\\
          \hline
          avg enc\_runtime (s)&1.05& 1.06& 1.11& 0.92&0.86 &1.214&1.56&2.5\\
          \hline
          max enc\_runtime (s)&1.51& 1.92& 1.56&1.31& 1.32 &1.5&2.03&3.42\\
          \hline
       \end{tabular}}
       
        \caption{Evaluating the scalability of the CNF encoding.
        }
        \label{tabencsize}
\end{table*}

\subsection{Results}
We report the following results by setting the following parameters $nb\_trees =10$ and $max\_depth=24$ for the random forest classifier trained on the vicinity of an input sample $x$ as the surrogate model. 
The experiments were conducted on an average of 1500 instances picked randomly from the MNIST database. The predictions are made using \footnote{the results for the other digits are similar but can not be reported here because of space limitation} the "one-vs-all" BNN classifiers trained to recognize the 0,2,5,6 and 8 digits. 
Due to the limited number of pages, we only present the results for radius 250 with an average of 200 neighbors around $x$ for MNIST. As for the rest, we consider all instances as neighbors (radius equal to the number of features). 

\subsubsection*{Evaluating the CNF encoding feasibility}
We report our results regarding the size of the generated CNF formulae. We use the Tseitin Transformation \cite{tseitin1983complexity} to encode the propositional formulae into an equisatisfiable CNF formulae. 

\noindent \textbf{Table \ref{tabencsize}} shows that the generated random forest classifiers provide interesting results in term of fidelity (high accuracy of the surrogate models) and tractability (size of the CNF encoding).In Table \ref{tabencsize}, the size of CNF is expressed as {\it number of variables/number clauses}.
 We can see that the number of variables and clauses of CNF formulae remains reasonable and easily handled by the current SAT-solvers which confirms the feasibility of the approach. 
 
\subsubsection*{Evaluating the enumeration of symbolic explanations}
The objective here is to assess the practical feasibility of the enumeration (scalability) of $SR_x$ and $CF_x$ explanations. 
For the enumeration of $CF_x$, we use the \textit{ EnumELSRMRCache tool}\footnote{available at \url{http://www.cril.univ-artois.fr/enumcs/}} implementing the boosting algorithm for MCSes enumeration proposed in \cite{gregoire2018boosting} with a timeout set to 600s. 
 As for the $SR_x$ explanations, their enumeration can be easily done by exploiting the minimal hitting set duality relationship between MUSes and MCSes. Due to the page limitation, we only present the results about the enumeration of $CF_x$, but the results in terms of the number of explanations generated remain of the same order of magnitude.
 
\noindent We observe within \textbf{Table \ref{tabenum}} that the average run-time remains reasonable (note that the times shown in Table \ref{tabenum} relate to the time taken to list all the explanations. The solver starts to find the first explanations very promptly) and that the approach is efficient in practice for medium size BNN classifiers (as shown in the experiments for BNNs with around 800 variables). 
We also observe that the number of $CF_x$ may be challenging for a user to understand, hence the need for scoring them to filter them out and find the ones with the strongest influence on the prediction.

\begin{table*}[t]
\scriptsize
       \resizebox{\textwidth}{!}{ \begin{tabular}{p{1.75cm}|p{1.cm}|p{1.cm}|p{1.cm}|p{1.cm}|p{1.cm}|p{1.cm}|p{1.cm}|p{1.cm}|p{1.cm}|}
        \hline
          & MNIST\_0&MNIST\_2  & MNIST\_5 & MNIST\_6& MNIST\_8 & SPECT & MONKS&Breast\_cancer \\  
          \hline
             min $\#$CFs &10&13&10&15&6 &15&3&11\\
          \hline
         avg $\#$CFs &35790&63916&99174&79520&4846 &204&15&947\\
          \hline
          max $\#$CFs &285219&546005&633416&640868&65554 &700&41&5541\\
          \hline
          min enumtime (s)&0.005&0.11&0.006& 0.11&0.008 &0.01&0.01&0.02 \\
          \hline
          avg enumtime (s)&21.49&42.11&77.72&50.86&2.35 &0.12&0.03&1.5\\
          \hline
          max enumtime (s)&234.18&600&600&531.16&35.08  &0.42&0.06&10.7\\
          \hline
       \end{tabular}}
      
        \caption{Evaluating the enumeration of counterfactual explanations.
        }
         \label{tabenum}
\end{table*}
\subsubsection*{Illustrating $SR_x$ and  $CF_x$ explanations for MNIST data set}
We trained two "one-vs-all"\footnote{A "one-vs-all" BNN $f_i$ returns a positive prediction for an input image representing the "i" digit, and negative one otherwise.} BNNs $f_8$ and $f_0$ to recognize the eight and zero digits. They have respectively achieved an accuracy of $97\%$ and $99\%$. The "a" column in the different figures shows the input images (resp. representing the digit 5 and 1). Those
data samples were negatively predicted. The model $f_8$ (resp. $f_0$) recognizes that the input image in the $1^{st}$ line (resp. the $2^{nd}$) is not an 8-digit (resp. a 0-digit). 
\begin{figure*}[t]\vspace{-0.1cm}
\begin{subfigure}{.5\textwidth}
  \centering
  \includegraphics[width=.75\linewidth]{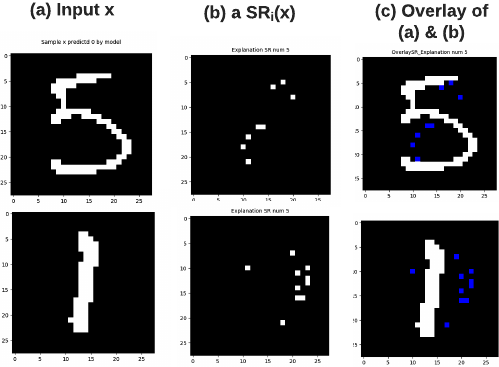}  
  \caption{Example of $SR_x$ explanations.}
  \label{fig:sub-first}
\end{subfigure}
\begin{subfigure}{.5\textwidth}
  \centering
  \includegraphics[width=.75\linewidth]{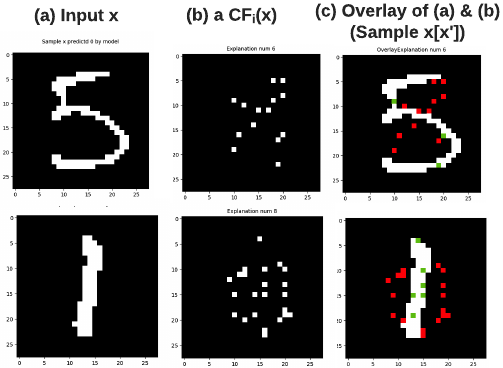}  
  \caption{Example of $CF_x$  explanations.}
  \label{fig:sub-second}
\end{subfigure}
\caption{Data samples from MNIST database and their respective symbolic explanations.}
\label{xpimgs}
\end{figure*}
Figure \ref{fig:sub-first} shows an example of a single  $SR_x$ explanation highlighting the sufficient pixels for the models $f_8$ and $f_0$ to trigger a negative prediction. Figure \ref{fig:sub-second} shows an example of $CF_x$ explanations showing the pixels to invert in the input images to make the models $f_8$ and $f_0$ predict them positively. 
 In addition, one could recognize in the "c" column of Fig. \ref{fig:sub-second} a pattern of the 8-digit for the first image, and 0 for the second. It gives us a kind of "pattern/template" of the images that the model would positively predict. \\
Figure \ref{figscore} shows heatmaps corresponding to the \textit{Feature Involvement (FI)} scores (column "b-c") and \textit{Feature Responsibility (FR)} (column "d-e") scores of the different input variables implicated in the $SR_x$ and $CF_x$. Visually, they are simpler, clearer and easier to understand and use. We used around 100 data samples to compare the most important features according to the $\mathcal{FI}$ score of our approach and those of SHAP ("f" column of Fig. \ref{figscore}). The results coincide from 20\% to 46\% of cases, which is visually confirmed in our figures.\vspace{-0.1cm}

\begin{figure*}[b]
   \includegraphics[width=0.75\textwidth]{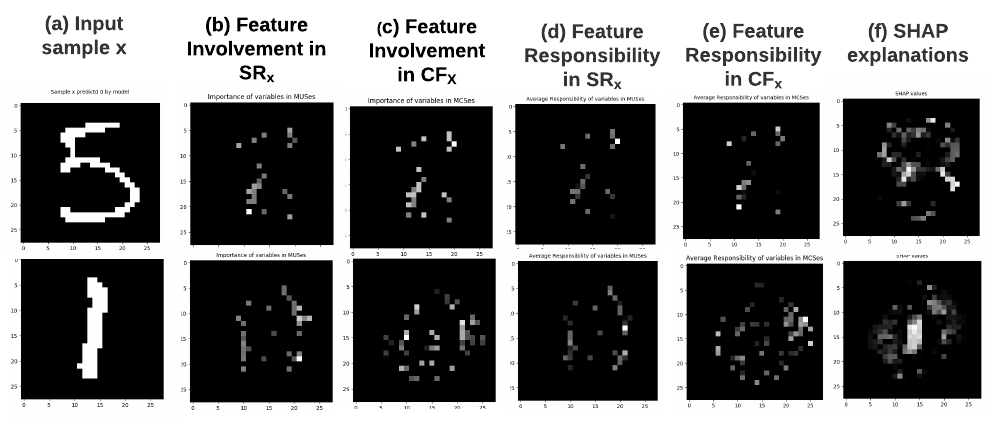}
\caption{Heatmaps in columns (b-c) representing the (FI) score, and (d-e) the (FR) computed over the $SR_x$ and $CF_x$ of the samples data from MNIST (column a) in comparaison to heatmaps of the SHAP values (column f).}
\label{figscore}
\end{figure*}

\section{Related Works}
Explaining machine learning systems has been a hot research topic recently.
There has been hundreds of papers on ML explainability but we will be focusing on the ones closely related to our work.
In the context of model-agnostic explainers where  the learning
function of the input model and its parameters are not known (black-box), we can cite some post-hoc explanations methods such as: LIME (Local Interpretable Model-agnostic Explanations) \cite{ribeiro2016should} which explain black-box classifiers by training an interpretable model $g$ on samples randomly generated in the vicinity of the data instance. We follow an approach similar to LIME, the difference is that we encode our surrogate model into a CNF to generate symbolic explanations. 
The authors in \cite{ribeiro2018anchors} proposed a High-precision model agnostic explanations called ANCHOR. It is based on computing a decision rule linking the feature space to a certain outcome, and consider it as an anchor covering similar instances. Something similar is done in SHAP (SHapley Additive exPlanations) \cite{lundberg2017unified} that provides explanations in the form of the game theoretically optimal called Shapley values. Due to its computational complexity, other model-specific versions have been proposed for linear models and deep neural networks (resp LinearSHAP and DeepSHAP) in \cite{lundberg2017unified}.
The main difference with this rule sets/feature-based explanation methods and the symbolic explanations we propose is that ours associates a score w.r.t to some relevance properties, in order to assess to what extent the measured entity is relevant as explanation or involved as features in the sufficient reasons or in the counterfactuals. 

Recently, some authors propose symbolic and logic-based XAI approaches that can be used for different purposes \cite{darwiche2020}. We can distinguish the compilation-based approaches where Boolean decision functions of classifiers are compiled into some symbolic forms. For instance, in \cite{chan2012reasoning,shih2018symbolic} the authors showed how to compile the decision functions of naive Bayes classifiers into a symbolic representation, known as Ordered Decision Diagrams (ODDs).
{We proposed in a previous work \cite{boumazouza2020symbolic} an approach designed to equip such symbolic approaches \cite{shih2018symbolic} with a module for counterfactual explainability.
There are some ML models whose direct encoding into CNF is possible. For instance, the authors in \cite{narodytska2018verifying} proposed a CNF encoding for Binarized Neural Networks (BNNs) for verification purposes. In \cite{shi2020tractable}, the authors propose a compilation algorithm of BNNs into tractable representations such as Ordered Binary Decision Diagrams (OBDDs) and Sentential Decision Diagrams (SDDs).  The authors in \cite{shih2019compiling} proposed  algorithms for compiling Naive and Latent-Tree Bayesian network classifiers into decision graphs. In \cite{KR2020-86}, the authors dealt with  a set of explanation queries and their computational complexity once classifiers are represented with compiled representations. 
However, the compilation-based approaches are hardly applicable to large sized models, and remain strongly dependent on the type of classifier to explain (non agnostic). Our approach can use those compilation algorithms to represent the whole classifier when the encoding remains tractable, but in addition, we propose a local
approximation of the original model using a surrogate model built on the neighborhood of the instance at hand. 

Recent works in \cite{ignatiev2019relating,ignatiev2019abduction}  deal with some forms  of symbolic explanations referred to as abductive explanations (AXp) and contrastive explanations (CXp) using SMT oracles. In \cite{ignatiev2021sat}, the authors explain the prediction of  decision list classifiers using a SAT-based approach. Explaining random forests and decision trees is dealt with for instance in \cite{KR2020-86} and \cite{DBLP:journals/corr/abs-2012-11067,DBLP:journals/corr/abs-2010-11034} respectively.
The main difference with our work, is that we are proposing an approach that goes from the model whose predictions are to be explained to its encoding 
and goes beyond the enumeration of symbolic explanations by defining some scoring functions w.r.t some relevance properties. 
Different explanation scores have been proposed in the literature. Authors in \cite{DBLP:journals/corr/abs-2011-07423} used the counterfactual explanations to define an explanation responsibility score for a feature value in the input. In \cite{DBLP:conf/sum/Bertossi20}, the authors used the answer-set programming to analyze and reason about diverse alternative counterfactuals and to investigate the causal explanations and the responsibility score in databases. 

\section{Concluding remarks and Discussions}
We proposed a novel model agnostic generic approach to explain individual outcomes by providing two complementary types of symbolic explanations (\textit{sufficient reasons} and \textit{counterfactuals}) and scores-based ones. The objective of the approach is to explain the predictions of a  black-box model by providing both symbolic and score-based explanations 
with the help of Boolean satisfiability concepts. The approach takes advantage of the strengths of already existing and proven solutions, and of the powerful practical tools for the generation of  MCS/MUS. 
The proposed approach overcomes the complexity of encoding a ML classifier into an equivalent logical representation by means of a surrogate model to symbolically approximate the original model in the vicinity of the sample of interest. The presentation of the paper was limited to the explanation of negative predictions to exploit the concepts of MUS and MCS and use a SAT-based approach. For positively predicted instances, we can simply work on the negation of the symbolic representation (CNF) of $f$ (namely $\neg \Sigma_f$). The enumeration of the explanations is done in the same  way as for negative predictions.\\
To the best of our  knowledge, our approach is the first that generates different types of symbolic explanations and  {\bf fine-grained} score-based ones. In addition, our approach is {\bf agnostic} and {\bf declarative}. 
 Another advantage of our approach is the {\bf local faithfulness} \cite{ribeiro2016should} to the instance to be explained.
 As future works, we intend to extend our approach for multi-label (ML) classification tasks to explain predictions in a multi-label setting. 


\section*{Acknowledgments} 
The authors would like to thank the Région Hauts-de-France and the University of Artois for supporting this work. 
\bibliographystyle{ACM-Reference-Format}
\bibliography{mybibliography}

\end{document}